\documentclass{article}

\usepackage[final]{neurips_2019}  

\usepackage[utf8]{inputenc}
\usepackage[T1]{fontenc}
\usepackage{hyperref}
\usepackage{url}
\usepackage{booktabs} 
\usepackage{amsfonts} 
\usepackage{nicefrac} 
\usepackage{microtype}

\usepackage{amsmath,amssymb,amsfonts}
\usepackage{subcaption}
\usepackage{bm}
\usepackage[dvipdfmx]{graphicx}
\usepackage{color}
\usepackage{tabularx}
\usepackage{algorithm}
\usepackage[noend]{algpseudocode}

\title{Simple and Scalable\\ Parallelized Bayesian Optimization}

\author{%
  Masahiro Nomura \\
  CyberAgent, Inc. \\
  \texttt{nomura\_masahiro@cyberagent.co.jp} \\
}

\begin{document}

\maketitle

\begin{abstract}
In recent years, leveraging parallel and distributed computational resources has become essential to solve problems of high computational cost.
Bayesian optimization (BO) has shown attractive results in those expensive-to-evaluate problems such as hyperparameter optimization of machine learning algorithms.
While many parallel BO methods have been developed to search efficiently utilizing these computational resources, these methods assumed synchronous settings or were not scalable.
In this paper, we propose a simple and scalable BO method for asynchronous parallel settings.
Experiments are carried out with a benchmark function and hyperparameter optimization of multi-layer perceptrons, which demonstrate the promising performance of the proposed method.
\end{abstract}

\section{Introduction}
\label{sec:intro}
Black-box optimization is a significant problem that appears in the fields of science and industry.
In black-box optimization, no algebraic representation of the objective function is given, and no gradient information is available; we can only observe the evaluation value for the sampled point.

One of the prominent methods in black-box optimization is Bayesian optimization (BO)~\citep{frazier2018tutorial,shahriari2015taking,brochu2010tutorial}.
There are many successful examples of BO for problems such as hyperparameter optimization of machine learning algorithms~\citep{swersky2013mtbo,snoek2012practical,bergstra2011algorithms,feurer2019hyperparameter} and neural architecture search~\citep{kandasamy2018neural,liu2018progressive}.
BO has high efficiency by using the observed data and dealing with the exploration-exploitation tradeoff.

In today's world, when computer infrastructure is becoming easier to prepare by cloud computing services, it is progressively vital to evaluate the objective function in parallel.
In sequential optimization, the next search cannot be performed until an evaluation of a previous point finished. By contrast, using parallel optimization allows multiple points to be evaluated at the same time, thus increasing efficiency depending on the resource.
The expansion of BO to parallel environments has been actively studied~\citep{wang2017batched,kathuria2016batched,wu2016parallel,shah2015parallel,desautels2014parallelizing,contal2013parallel,kandasamy2018parallelised,ginsbourger2011dealing,janusevskis2012expected,wang2016parallel}.
While the majority of these studies in parallel BO assume batch evaluations (i.e., synchronous settings), this is a problem when the computational cost varies greatly by evaluation point.
For example, in hyperparameter optimization, the training time of a model varies greatly depending on the hyperparameter configuration.
Several BO methods for asynchronous parallel settings exist~\citep{kandasamy2018parallelised,ginsbourger2011dealing,janusevskis2012expected,wang2016parallel,snoek2012practical}.
However, the time complexity of these methods is cubic with respect to the number of points evaluated; this can be an obstacle when applied to more massive scale problems.

In this paper, we propose a simple and scalable BO method for asynchronous parallel settings.
The proposed method inherits the scalability of the Tree-structured Parzen Estimator~\citep{bergstra2011algorithms}, whose time complexity is linear.
In our experiments for a benchmark function and hyperparameter optimization of multi-layer perceptrons, we demonstrate that the proposed method achieves faster convergence within a fixed evaluation budget than the asynchronous parallel Thompson Sampling~\citep{kandasamy2018parallelised}, which is one of the state-of-the-art methods in parallel BO.

\section{Background}
\label{sec:bg}

\subsection{Bayesian Optimization}
Bayesian optimization (BO) is an efficient method for black-box optimization, which is performed by repeating the following steps:
(1) Based on the data observed thus far, BO constructs a surrogate model that considers the uncertainty of the objective function.
(2) BO calculates the acquisition function to determine the point to be evaluated next by using the surrogate model.
(3) By utilizing the acquisition function, BO determines the next point to evaluate.
(4) BO then updates the surrogate model based on the newly obtained data, then returns to step (2).

Popular acquisition functions in BO include the expected improvement (EI)~\citep{jones1998efficient}, probability of improvement~\citep{kushner1964new}, upper confidence bound~\citep{srinivas2009gaussian}, mutual information~\citep{contal2014gaussian}, and knowledge gradient~\citep{frazier2009knowledge}.
Among them, EI is one of the most common acquisition functions in BO and calculated as follows:
\begin{align}
    \label{eq:ei}
    a_{\rm EI}({\bf x}) = \int_{-\infty}^{\infty} \max(y^{*} - y, 0) p(y | {\bf x}) dy,
\end{align}
where $y^{*}$ is some threshold.
In the Gaussian process-based BO~\citep{snoek2012practical}, the best evaluation value so far is typically used for $y^{*}$.

\subsection{Tree-structured Parzen Estimator}
The Tree-structured Parzen Estimator (TPE)~\citep{bergstra2011algorithms,bergstra2013making} is one of the most popular BO methods and widely used in hyperparameter optimization~\citep{bergstra2013hyperopt,eggensperger2013towards,falkner2018bohb,akiba2019optuna}.
In the calculation of EI in Equation (\ref{eq:ei}), TPE models $p({\bf x} | y)$ and $p(y)$ using the following equations instead of modeling $p(y | {\bf x})$:
\small
\begin{align}
    \label{eq:tpe_p_x_y}
    &p({\bf x} | y) =\begin{cases}
    l({\bf x}) & (y < y^{*}) \\
    g({\bf x}) & (y \geq y^{*}),
    \end{cases}\\
    \label{eq:tpe_p_y}
    &p(y < y^{*}) = \gamma.
\end{align}
\normalsize
Using the Equation (\ref{eq:tpe_p_x_y}) and (\ref{eq:tpe_p_y}), TPE calculates EI as follows:
\small
\begin{align}
    a_{\rm EI}({\bf x}) &\propto \Bigl( \gamma + \frac{g({\bf x})}{l({\bf x})} (1 - \gamma) \Bigr)^{-1}.
\end{align}
\normalsize
In each iteration, TPE selects ${\bf x}$ that maximizes $a_{\rm EI}({\bf x})$ as the next evaluation point; that is, TPE selects  ${\bf x}$ that maximizes $l({\bf x}) / g({\bf x})$.
TPE uses a kernel density estimator to model $l({\bf x})$ and $g({\bf x})$.
Compared to the Gaussian process-based BO~\citep{snoek2012practical}, whose time complexity is cubic for the number of points evaluated, the time complexity of TPE is linear and therefore scalable.

\subsection{Parallelized Bayesian Optimization}
Although there are abundant parallel BO methods~\citep{wang2017batched,kathuria2016batched,wu2016parallel,shah2015parallel,desautels2014parallelizing,contal2013parallel,kandasamy2018parallelised,ginsbourger2011dealing,janusevskis2012expected,wang2016parallel},  many of those methods assume synchronous settings.
For problems such as hyperparameter optimization where the computation time of the evaluation can vary greatly depending on the evaluation point, the optimization method should be adapted to the asynchronous setting.
There are several BO methods for asynchronous settings \citep{kandasamy2018parallelised,ginsbourger2011dealing,janusevskis2012expected,wang2016parallel,snoek2012practical}.
However, these methods have a problem of scalability because the time complexity is cubic.
On the other hand, the proposed method inherits the advantages of TPE and enables optimization with linear time complexity.

\section{Proposed Method}
We propose a new optimization method that is simple and scalable for asynchronous parallel settings.
The proposed method tries to sample more points for regions with a high probability of having an evaluation value better than a threshold $y^{*}$; in other words, the points are sampled from the distribution $p(y < y^{*} | {\bf x})$.
We note that, unlike typical Bayesian optimization, which selects ${\bf x}$ that optimizes the acquisition function, the proposed method samples the points directly from the distribution $p(y < y^{*} | {\bf x})$.
By using Equation (\ref{eq:tpe_p_x_y}), (\ref{eq:tpe_p_y}), and $p({\bf x}) = \int_{\mathbb{R}} p({\bf x} | y) p(y) dy = \gamma l({\bf x}) + (1-\gamma) g({\bf x})$, $p(y < y^{*} | {\bf x})$ is calculated as follows:
\small
\begin{align}
\label{eq:main}
\begin{aligned}
    p(y < y^{*} | {\bf x}) &= \int_{-\infty}^{y^{*}} p(y | {\bf x}) dy
    = \int_{-\infty}^{y^{*}} \frac{p({\bf x} | y) p(y)}{p({\bf x})} dy
    = \frac{\gamma l({\bf x})}{\gamma l({\bf x}) + (1 - \gamma) g({\bf x})}.
\end{aligned}
\end{align}
\normalsize

\begin{algorithm}[t]
\caption{Proposed Method in asynchronous parallel settings}
\label{alg:proposed}
\begin{algorithmic}[1]
    \Require objective function $f$, quantile $\gamma$
    \State $\mathcal{D}_1 = \varnothing$
    \For{$j = 1, 2, \cdots$} \Comment{\texttt{every time one of workers become free}}
        \State Sample a point ${\bf x}_j$ from the distribution $p(y < y^{*} | {\bf x}_j)$ using rejection sampling
        \State $y_j =$ evaluate $f({\bf x}_j)$
        \State $\mathcal{D}_{j+1} = \mathcal{D}_{j} \cup \{ ({\bf x}_j, y_j) \}$
        \State Compute $p(y < y^{*} | {\bf x})$ by using $\gamma$ and $\mathcal{D}_{j+1}$ \Comment{\texttt{in Equation(\ref{eq:main})}}
    \EndFor
\end{algorithmic}
\end{algorithm}

Algorithm \ref{alg:proposed} shows the overall algorithm of the proposed method in the asynchronous setting.
Because the proposed method samples points probabilistically, it can be naturally extended to asynchronous parallel settings.
We sample from the distribution $p(y < y^{*} | {\bf x})$ using rejection sampling.
In the proposed method, one-dimensional kernel density estimation is performed hierarchically to obtain $l({\bf x})$ and $g({\bf x})$, as in \citep{bergstra2011algorithms,bergstra2013making}.

While the proposed method is inspired by TPE in modeling $p({\bf x} | y)$ and $p(y < y^{*}) = \gamma$, the attitude of these methods to the parallel setting is different.
TPE always selects ${\bf x}$ that maximizes $l({\bf x}) / g({\bf x})$; therefore, it continues to select the same or similar points for all workers in parallel settings.
On the other hand, in the proposed method, the above situation can be avoided because the points are sampled probabilistically.

\section{Experiments}
In this section, we assess the performance of the proposed method through a benchmark function and the hyperparameter optimization of machine learning algorithms.
We compared the proposed method with the asynchronous parallel Thompson Sampling (Parallel-TS) \citep{kandasamy2018parallelised}, which is a Bayesian optimization method using a Gaussian process as a surrogate model and Thompson Sampling as an acquisition function.
We do not compare with other asynchronous methods such as asynchronous random sampling, asynchronous version of upper confidence bound~\citep{desautels2014parallelizing,ginsbourger2011dealing}, and asynchronous expected improvement~\citep{jones1998efficient} because Parallel-TS shows better performance than these methods~\citep{kandasamy2018parallelised}.
For each method, the evaluation budget for optimization was set to $500$ seconds.
We set the number of parallel workers to $4$ and performed $10$ trials for each problem.
Details of the experimental settings are shown in Appendix \ref{appendix:exp_setup}.

In the first experiment, we optimized the Hartmann18 function, which is 18-dimensional, used in \citep{kandasamy2018parallelised}.
Hartmann18 function was constructed as $f({\bf x}_{1:18}) = h({\bf x}_{1:6}) + h({\bf x}_{7:12}) + h({\bf x}_{13:18})$ where $h$ is the six-dimensional Hartmann function~\citep{simulationlib}.
We modeled the evaluation time as a random variable that is drawn from a half-normal distribution~\citep{kandasamy2018parallelised}.
Further details of the experiment of the benchmark function is shown in Appendix \ref{appendix:exp_bench}.

In the second experiment, we conducted the hyperparameter optimization of multi-layer perceptrons (MLPs).
The MLP consists of two fully-connected layers with softmax at the end.
Table \ref{tab:hp_mlp} shows the six hyperparameter of MLP we optimized and their respective search spaces.
We regarded the integer-valued hyperparameters as continuous variables by using rounded integer when evaluating.
We set the maximum number of epochs during training to $20$, and the mini-batch size to $128$.
MLPs were trained on Fashion-MNIST clothing articles dataset~\citep{xiao2017fashion}.
We used $50,000$ images as training data and $10,000$ images as validation data.
The misclassification rate on the validation dataset was used as the evaluation value.

\begin{table}[t]
    \centering
    \caption{Details of hyperparameters of MLP on Fashion-MNIST dataset.}
     \label{tab:hp_mlp}
    \setlength\tabcolsep{2mm}
    \begin{tabularx}{\textwidth}{p{18em}p{5em}p{10em}}
    \toprule
         Hyperparameters & Type & Range \\ 
         \midrule
         Learning Rate                            & float & $[1.0 \times 10^{-3}, 0.2]$\\
         Momentum                                 & float & $[0.8,0.99]$\\
         Number of Hidden Nodes (1st Layer)                       & int & $[50, 500]$ \\
         Number of Hidden Nodes (2nd Layer)                      & int  & $[50, 500]$ \\
         Dropout Rate (1st Layer)                            & float & $[0,0.8]$ \\
         Dropout Rate (2nd Layer)                            & float & $[0,0.8]$ \\
         \bottomrule
    \end{tabularx}
\end{table}

\begin{figure*}[tb]
\centering
  \begin{minipage}[t]{.94\textwidth}
    \centering
    \includegraphics[width=110mm]{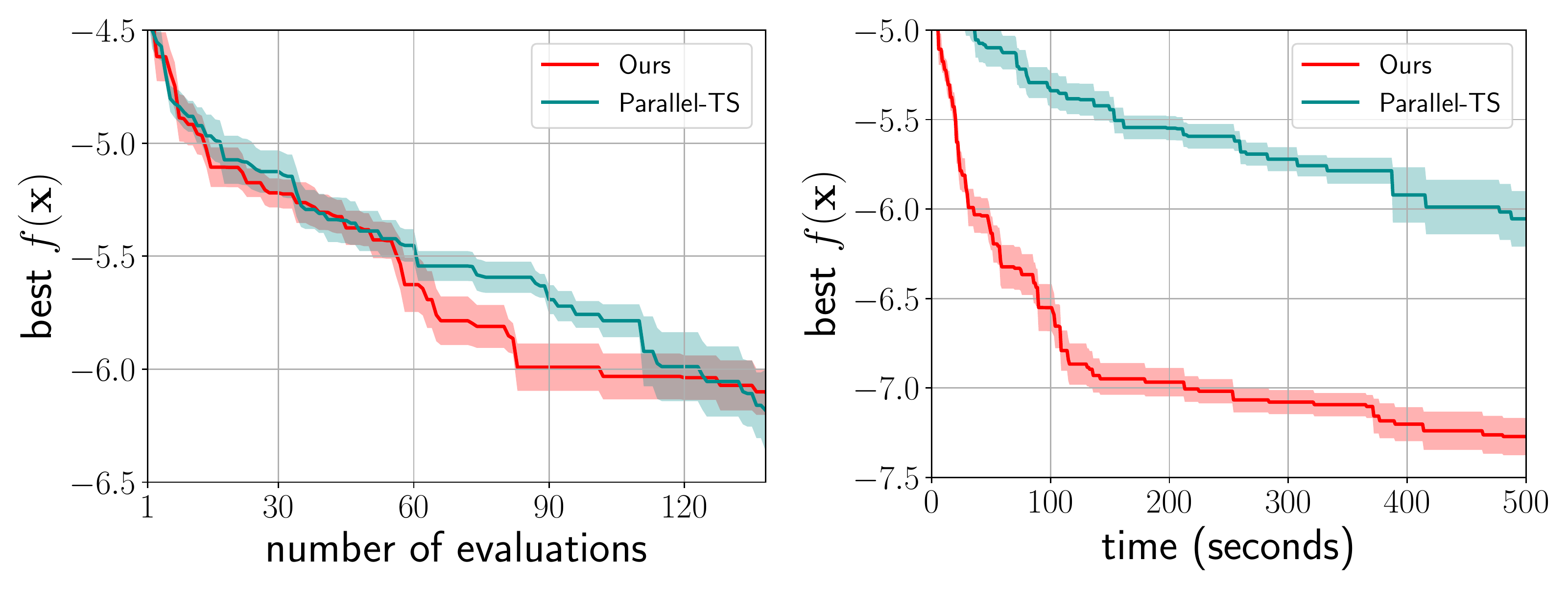}
    \subcaption{Hartmann18 function}
    \label{subfig:hartmann}
  \end{minipage}
  \begin{minipage}[t]{.94\textwidth}
    \centering
    \includegraphics[width=110mm]{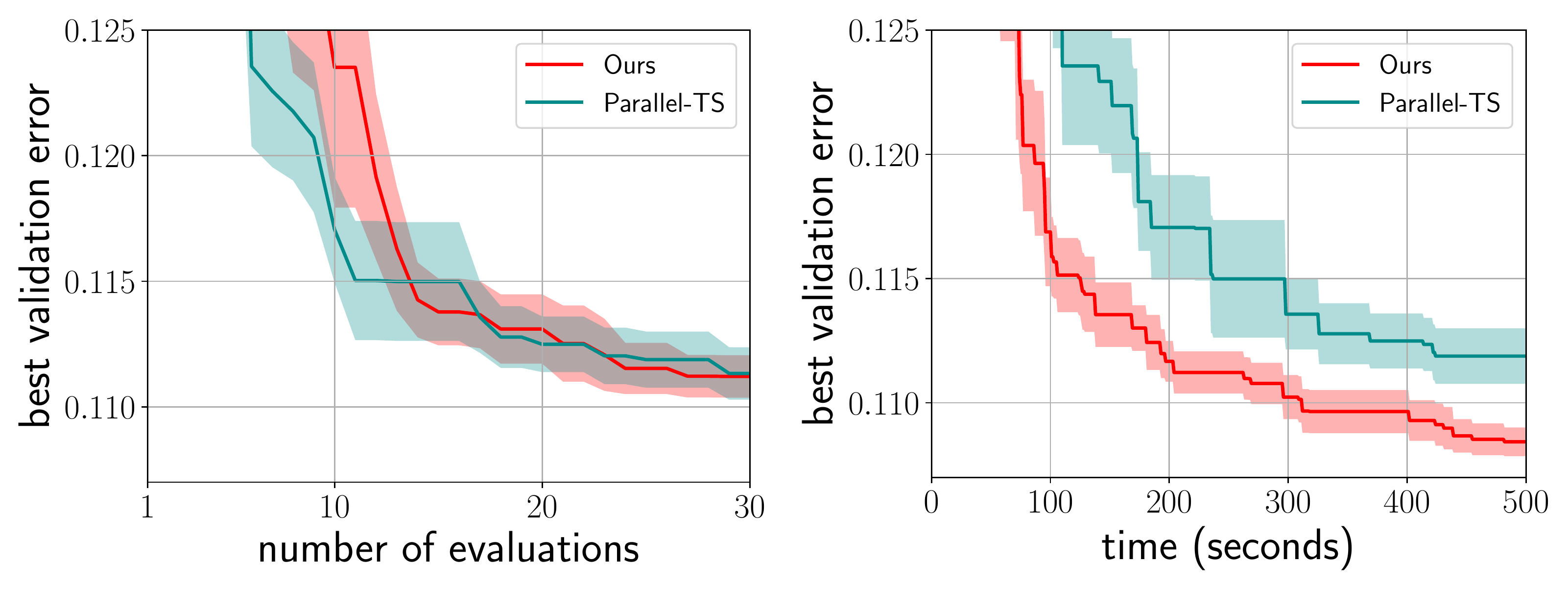}
    \subcaption{MLP on Fashion-MNIST}
    \label{subfig:fashion_mnist_mlp}
  \end{minipage}
  \caption{The sequences of the mean and standard error of the best evaluation values on the benchmark function and hyperparameter optimization. The x-axis denotes (left) the number of evaluations or (right) the computation time.}
  \label{fig:exp_bench_and_hpo}
\end{figure*}

The results of these experiments are shown in Figure \ref{fig:exp_bench_and_hpo}.
To assess the search efficiency and calculation cost of the proposed method separately, we show two figures;
in each problem, the figure on the left shows the number of evaluations on the x-axis, and the figure on the right shows the calculation time on the x-axis.
In both of the problems, the performance of the proposed method shows competitive results with that of Parallel-TS in terms of the number of evaluations; on the other hand, in terms of computation time, the proposed method outperforms Parallel-TS.
This result is because Parallel-TS needs to sample many points that follow the posterior distribution of the Gaussian process.
In contrast, the proposed method does not require such expensive calculations.

\section{Conclusion}
\label{sec:conclusion}
In this paper, we proposed a simple and scalable parallelized Bayesian optimization method for black-box optimization.
The proposed method enables efficient search by generating more points in regions that are likely to be promising.
By inheriting the useful properties of the Tree-structured Parzen Estimator (TPE)~\citep{bergstra2011algorithms}, the proposed method achieves linear time complexity.
Furthermore, the proposed method can be naturally extended to asynchronous settings because it is a probabilistic sampling method.
The experiments on the benchmark function and hyperparameter optimization have shown the effectiveness of the proposed method.

As future work, we would like to confirm whether the performance of BOHB~\citep{falkner2018bohb} will improve when the TPE part of BOHB, which combines Hyperband and TPE, is replaced with the proposed method.
BOHB deals with the parallelization in an ad-hoc way which reduces the number of samples used to optimize the acquisition function in TPE.
We believe that the performance of BOHB may be improved by using the more motivated proposed method.

\subsubsection*{Acknowledgments}
The author would like to thank Masashi Shibata for his valuable advice with the experiments.

\section*{References}
\medskip

\small

\begingroup
\renewcommand{\section}[2]{}%

\bibliography{main}
\bibliographystyle{iclr2020_conference}

\endgroup

\clearpage

\appendix

\section{Details of Experimental Setup}
\label{appendix:exp_setup}
For the asynchronous parallel Thompson Sampling (Parallel-TS) , we used a squared exponential kernel with the bandwidth for each dimension, the scale parameter of the kernel, and the noise variance, as stated in \citep{kandasamy2018parallelised}.
These kernel hyperparameters were updated by maximizing the data likelihood after each iteration.
We sampled $N_{\rm acq}$ points from the posterior distribution of the Gaussian process and selected the ${\bf x}$ with the smallest value as the next point to evaluate.
Using the authors' repository\footnote{https://github.com/kirthevasank/gp-parallel-ts} of Parallel-TS as a reference, we set $N_{\rm acq} = \lfloor \max (2000, \min(5, d) \cdot \sqrt{\min(j, 1000)}) \rfloor$, where $d$ is the dimension, and $j$ is the number of completed evaluations.

We used the kernel density estimator implemented in Optuna~\citep{akiba2019optuna} to construct $l({\bf x})$ and $g({\bf x})$ in the proposed method.
We set the quantile parameter in Equation (\ref{eq:main}) to $\gamma = 0.1$ through the experiments.

\section{Details of Benchmark Function}
\label{appendix:exp_bench}
In the experiment of the Hartmann18 function, we model the evaluation time as a random variable that is drawn from a half-normal distribution~\citep{kandasamy2018parallelised}.
The probability density function of the half-normal is given by
\begin{align}
    p(x; \sigma) = \frac{\sqrt{2}}{\sigma \sqrt{\pi}} \exp \Bigl( -\frac{x^2}{2\sigma^2} \Bigr),
\end{align}
where $\sigma$ is a scale parameter.
Because the expectation of the random variable that follows this distribution is $\sigma \sqrt{2} / \sqrt{\pi}$, we set $\sigma = \sqrt{\pi / 2}$ so that the expectation is $1$.



\end{document}